\documentclass[10pt,twocolumn,letterpaper]{article}

\usepackage[pagenumbers]{cvpr}   

\usepackage{booktabs}
\usepackage{tabularx}
\usepackage{array}
\usepackage{makecell}
\usepackage{amsmath}

\definecolor{cvprblue}{rgb}{0.21,0.49,0.74}
\usepackage[pagebackref,breaklinks,colorlinks,allcolors=cvprblue]{hyperref}

\def\paperID{*****}
\def\confName{CVPR}
\def\confYear{2026}

\newcolumntype{L}[1]{>{\raggedright\arraybackslash}p{#1}}
\newcolumntype{Y}{>{\raggedright\arraybackslash}X}

\title{Beyond Model Design: Data-Centric Training and Self-Ensemble for Gaussian Color Image Denoising}

\author{
Gengjia Chang$^{1}$ \quad Xining Ge$^{2}$ \quad Weijun Yuan$^{3}$ \quad Zhan Li$^{3}$\\
Qiurong Song$^{4}$ \quad Luen Zhu$^{4}$ \quad Shuhong Liu$^{5,\dagger}$\\
$^{1}$Hefei University of Technology \quad
$^{2}$Hangzhou Dianzi University \quad
$^{3}$Jinan University\\
$^{4}$South China Agricultural University \quad
$^{5}$The University of Tokyo
}

\begin{document}
\maketitle
\begingroup
\renewcommand{\thefootnote}{}
\renewcommand{\theHfootnote}{challenge-note}
\footnotetext{The \emph{NTIRE 2026 Image Denoising Challenge (noise level=50)} is hosted on Codabench. Official page: \url{https://www.codabench.org/competitions/12905/}.}
\addtocounter{footnote}{-1}
\endgroup

\begin{abstract}
This paper presents our solution to the NTIRE 2026 Image Denoising Challenge (Gaussian color image denoising at fixed noise level $\sigma = 50$). Rather than proposing a new restoration backbone, we revisit the performance boundary of the mature Restormer architecture from two complementary directions: stronger data-centric training and more complete test-time enhancement. Starting from the public Restormer $\sigma\!=\!50$ baseline, we expand the standard multi-dataset training recipe with larger and more diverse public image datasets and organize optimization into two stages. At inference, we apply $\times 8$ geometric self-ensemble to further extract additional performance from the trained model. A TLC-style local inference wrapper is retained for implementation consistency; however, systematic ablation reveals its quantitative contribution to be negligible in this setting. On the challenge validation set of 100 images, our final submission achieves 30.762 dB PSNR and 0.861 SSIM, improving over the public Restormer $\sigma\!=\!50$ pretrained baseline by up to 3.366 dB PSNR. Ablation studies show that the dominant gain originates from the expanded training corpus and the two-stage optimization schedule, and self-ensemble provides marginal but consistent improvement.
\end{abstract}

\section{Introduction}
Image denoising is a fundamental low-level vision task that underpins a broad spectrum of downstream applications, including image restoration~\cite{ge2026clip,ren2026esr}, autonomous driving~\cite{lisgs2025,liumg2025,zhou2024mod}, VR/AR~\cite{lidense2025}, 3D reconstruction~\cite{liu2025i2nerf,cui2026unifying,liu20263drr}, scene understanding~\cite{liuderain2025,liu2025realx3d,liudenoise2026}, and scientific discovery~\cite{liudenoise2026}. For Gaussian color image denoising at a fixed noise level, deep learning has already established a mature technical spectrum. Representative CNN-based methods such as DnCNN~\cite{zhang2017beyond}, FFDNet~\cite{zhang2018ffdnet}, and RIDNet~\cite{anwar2019real} demonstrate that residual learning, noise-aware conditioning, and feature attention can substantially improve restoration quality. Multi-stage restoration frameworks such as MPRNet~\cite{zamir2021multi} further show that progressive refinement can be as consequential as raw backbone capacity.

Transformer-based approaches have pushed this frontier further. SwinIR~\cite{liang2021swinir}, Uformer~\cite{wang2022uformer}, and Restormer~\cite{zamir2022restormer} establish that long-range dependency modeling and efficient multi-scale feature interaction are highly effective for image restoration. Among these, Restormer has emerged as a particularly strong public baseline for high-resolution restoration and Gaussian denoising, owing to its favorable balance between global context modeling and computational practicality. Once a backbone of this caliber is in place, however, the remaining performance gap may no longer be primarily attributable to architectural limitations. Instead, the binding constraint may shift to how much prior information the training data can provide, and to what extent the inference strategy can fully exploit what a well-trained checkpoint already encodes.

Motivated by this observation, we investigate how far a mature Restormer denoiser can be pushed through two complementary and practically accessible levers. On the training side, we expand the training dataset beyond the standard public recipe and organize optimization into two successive stages, leading to better checkpoints from greater visual diversity and more stable convergence. On the inference side, we apply $\times 8$ geometric self-ensemble to aggregate predictions across all symmetries of the dihedral group, yielding a more stable and consistent final output. A TLC-style local inference wrapper is retained in our implementation for consistency with prior practice, but systematic ablation shows that its quantitative contribution in this setting is negligible. Our experimental protocol is therefore explicitly designed to isolate what genuinely drives performance from what merely persists in the engineering pipeline.

The main contributions of this work are as follows. First, we demonstrate that a Restormer-based Gaussian denoiser can be substantially improved without any architectural modification, through expansion of the training data and a two-stage training schedule alone. Second, we provide a systematic quantification of the gain attributable to $\times 8$ geometric self-ensemble at inference, along with its associated computational overhead. Third, through controlled comparison against the public Restormer $\sigma\!=\!50$ pretrained baseline under a unified 100-image evaluation protocol, we establish that the dominant source of improvement is the stronger trained checkpoint rather than any test-time wrapper.

We also report our results in the NTIRE 2026 Image Denoising Challenge (noise level $= 50$) \cite{ntire26denoising}, where we participated under the team name \textbf{wedream}. Our submission achieved 29.89 dB PSNR and 0.87 SSIM, ranking \textbf{2nd} on the final leaderboard. This result provides an independent external validation of the effectiveness of the proposed training recipe under competitive benchmark conditions.

\section{Related Work}
\subsection{Deep Image Denoising}
Before modern transformer-based restorers, image denoising had already progressed through a long line of prior-based and convolutional methods. Classical approaches such as Non-Local Means~\cite{buades2005non}, BM3D~\cite{dabov2007image}, and K-SVD dictionary learning~\cite{elad2006image} showed that carefully designed image priors could recover strong structure under additive noise, while trainable reaction-diffusion models~\cite{chen2016trainable} began to bridge model-based restoration and learned inference. Early deep restoration backbones such as MemNet~\cite{tai2017memnet} further confirmed that effective denoising depends not only on local filtering strength, but also on how well intermediate representations are propagated across depth.

Deep image denoising then evolved from residual CNN denoisers to more structured restoration systems. DnCNN~\cite{zhang2017beyond} established residual learning as a strong foundation for Gaussian denoising, while FFDNet~\cite{zhang2018ffdnet} introduced explicit noise-level conditioning so that one model could handle multiple noise strengths. RIDNet~\cite{anwar2019real} emphasized feature attention within a single-stage design, and MPRNet~\cite{zamir2021multi} demonstrated that multi-stage restoration can improve difficult image recovery tasks through progressive refinement. Real-image and blind denoising work further broadened this line: CBDNet~\cite{guo2019toward}, CycleISP~\cite{zamir2020cycleisp}, the MIRNet family~\cite{zamir2020learning,zamir2022learning}, deep boosting~\cite{chen2019real}, and practical blind denoisers built on data synthesis~\cite{zhang2023practical} all stress that realistic degradations and stronger restoration pipelines matter as much as nominal backbone size. Related formulations such as plug-and-play reconstruction with deep denoiser priors~\cite{zhang2021plug} also show how denoisers can serve as reusable priors beyond direct feed-forward prediction.

A second important branch studies learning without paired clean targets. Noise2Noise~\cite{lehtinen2018noise2noise}, Noise2Void~\cite{krull2019noise2void}, Noise2Self~\cite{batson2019noise2self}, Self2Self~\cite{quan2020self2self}, Neighbor2Neighbor~\cite{huang2021neighbor2neighbor}, and high-quality self-supervised denoising~\cite{laine2019high} reveal that denoising performance can be learned from noisy observations alone under suitable independence assumptions. Adjacent temporal settings such as burst and video denoising, represented by Deep Burst Denoising~\cite{godard2018deep}, ViDeNN~\cite{claus2019videnn}, and FastDVDnet~\cite{tassano2020fastdvdnet}, further highlight how extra observation diversity can compensate for harder restoration conditions.

\subsection{Transformer Image Restoration}
Transformer-style restoration has become a mainstream direction because long-range dependency modeling is naturally useful for image recovery. This line can be traced through generic image-processing transformers such as IPT~\cite{chen2021pre} and the hierarchical Swin Transformer backbone~\cite{liu2021swin}, which in turn motivated restoration-oriented designs such as SwinIR~\cite{liang2021swinir}, Uformer~\cite{wang2022uformer}, and Restormer~\cite{zamir2022restormer}. For the Gaussian denoising setting considered here, Restormer is already a very strong public baseline, which makes it a suitable backbone for asking a different question: how much additional performance can still be extracted without inventing a new backbone.

At the same time, the broader restoration literature continues to diversify around alternative backbone and aggregation choices. HiNet~\cite{chen2021hinet}, NAFNet-style simple baselines~\cite{chen2022simple}, MAXIM~\cite{tu2022maxim}, CAT~\cite{chen2022cross}, and TLC-style global information aggregation~\cite{chu2022improving} each emphasize different trade-offs among efficiency, local structure modeling, and long-range interaction. More recent state-space and attention variants such as MambaIR~\cite{guo2024mambair}, MambaIRv2~\cite{guo2025mambairv2}, ELAN~\cite{zhang2022efficient}, SRFormer~\cite{zhou2023srformer}, and activated-pixel transformer variants~\cite{chen2023activating} further illustrate how image restoration continues to borrow from the wider design space of modern vision backbones. 

Recent challenge-era super-resolution systems also show that measurable gains can still come from inference-side aggregation without redesigning the main restorer. Training-free strong-branch compensation ensembles and dual-branch infrared super-resolution pipelines both combine complementary branches with TLC-style conversion or $\times 8$ geometric self-ensemble to further strengthen reconstruction quality~\cite{chang2026training,ge2026dual}.

\subsection{Data-Centric Training and test-time Enhancement}
Large and diverse image corpora are increasingly important in image restoration. DIV2K~\cite{agustsson2017ntire} and Flickr2K~\cite{lim2017flickr2k} are widely used high-resolution image sources in supervised restoration, while BSD500~\cite{arbelaez2011bsd500} and the Waterloo Exploration Database~\cite{ma2017wed} contribute additional natural image diversity. The ESRGAN line and its OutdoorSceneTraining collection~\cite{wang2018recovering,wang2018esrgan}, as well as larger-scale or officially released high-resolution sources such as LSDIR~\cite{li2023lsdir}, DIV8K~\cite{gu2019div8k}, LIU4K-v2~\cite{liu2020liu4kv2}, and the NKUSR8K release distributed with the DiT4SR project~\cite{duan2025nkusr8k}, further broaden the texture coverage and spatial diversity available for restoration training. Similar data-centric observations also appear in the broader real-world and blind super-resolution literature, where realistic image sources and degradation diversity are repeatedly emphasized~\cite{chen2022real,liu2022blind}.

For denoising specifically, realistic data generation, noise modeling, and benchmark construction are equally important. SIDD~\cite{abdelhamed2018high}, DND~\cite{plotz2017benchmarking}, and RENOIR~\cite{anaya2018renoir} provide influential real-noise benchmarks, while challenge tracks such as NTIRE 2019, 2020, and 2023~\cite{abdelhamed2019ntire,abdelhamed2020ntire,li2023ntire} make progress easier to compare across methods. Complementary studies on camera noise and synthesis, including cross-channel noise modeling~\cite{nam2016holistic}, conditional noise flows~\cite{abdelhamed2019noise}, and unprocessing pipelines for learned raw denoising~\cite{brooks2019unprocessing}, reinforce the view that stronger restoration systems often emerge from stronger data assumptions as much as from stronger architectures.

Related evidence also appears in recent adverse-condition 3D restoration. Zheng \etal\ guide smoke-scene reconstruction with multimodal visual priors, ELoG-GS strengthens extreme low-light 3D recovery through luminance-guided enhancement, and SmokeGS-R uses physics-guided pseudo-clean supervision to better separate appearance restoration from geometry recovery~\cite{zheng20263d,liu2026elog,fu2026smokegs}. GenSmoke-GS and Dehaze-then-Splat adopt multi-stage generative restoration before or during 3D Gaussian Splatting optimization, while Naka-GS combines low-light correction with progressive point pruning for cleaner Gaussian initialization~\cite{cao2026gensmoke,chen2026dehaze,zhu2026naka}. Reliability-aware staged low-light Gaussian Splatting points in the same direction by further underscoring the value of staged processing and reliability-aware design under degraded illumination~\cite{guo2026reliability}. Although these methods target degraded 3D reconstruction rather than Gaussian image denoising, they support the broader claim that restoration quality often depends on data preparation, enhancement priors, and staged optimization as much as on backbone design.

Test-time enhancement is a complementary but distinct direction. Geometric self-ensemble is widely used in restoration practice, as averaging predictions across flip and transpose transforms can reduce prediction variance with no retraining. Strong restoration baselines such as SwinIR and SRFormer explicitly report this test-time strategy~\cite{liang2021swinir,zhou2023srformer}. TLC-style global information aggregation~\cite{chu2022improving} is another practical technique for improving full-image restoration behavior through local conversion. In this paper, the emphasis is not on proposing a new inference module, but on carefully disentangling the gains from stronger training, self-ensemble, and the retained TLC-style wrapper.

\begin{figure*}[!t]
    \centering
    \includegraphics[width=0.94\textwidth]{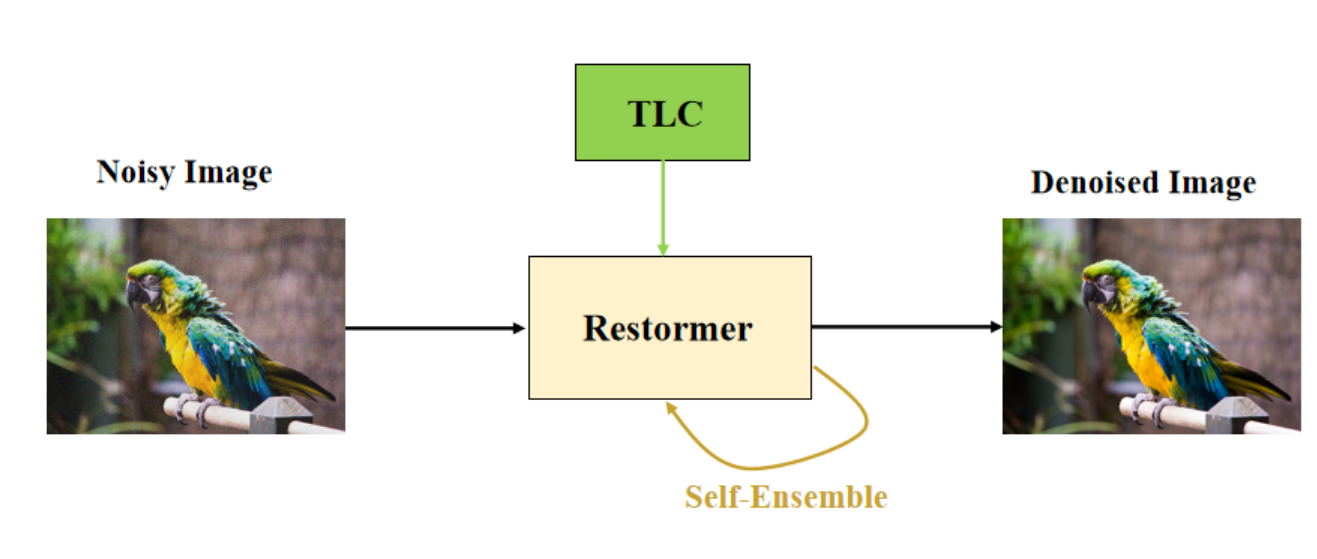}
    \caption{Overview of the final denoising pipeline. A noisy image is restored by Restormer, the TLC-style wrapper is retained for inference consistency, and $\times 8$ self-ensemble is used to refine the final denoised output.}
    \label{fig:pipeline}
\end{figure*}

\section{Method}
\subsection{Problem Definition and Framework Overview}
We consider Gaussian color image denoising at noise level $\sigma = 50$. Let $y$ denote the clean image and $x$ the observed noisy image. The degradation process is
\begin{equation}
x = y + n,\qquad n \sim \mathcal{N}(0, \sigma^2 I).
\end{equation}
Given $x$, the goal is to learn a restoration function $f_{\theta}(\cdot)$ that produces
\begin{equation}
\hat{y} = f_{\theta}(x),
\end{equation}
with $\hat{y}$ as close as possible to $y$.

As illustrated in Figure~\ref{fig:pipeline}, the overall framework has three components. The backbone is Restormer~\cite{zamir2022restormer}, adopted without architectural modification. The training pipeline strengthens the standard public recipe with a large-scale and diverse training dataset organized into a two-stage optimization schedule. The inference pipeline applies $\times 8$ geometric self-ensemble and retains a TLC-style local inference wrapper.

\subsection{Restormer Backbone}
Restormer~\cite{zamir2022restormer} adopts a hierarchical encoder-decoder structure. The input is first mapped to feature space through overlap patch embedding, then processed by multi-scale encoder stages that progressively enlarge the receptive field. The decoder restores spatial details through upsampling and skip connections, and a refinement module further improves the restored output. The final prediction is fused with the input through a global residual path.

Its core design combines multi-DConv head transposed attention with gated convolutional feed-forward blocks. This makes Restormer particularly suitable for high-resolution restoration because it preserves long-range context modeling while still keeping local structure sensitivity. Since the backbone is already strong, we do not alter it aggressively. Instead, the present paper concentrates on the training recipe and inference behavior wrapped around this backbone.

\subsection{Large-scale Training Dataset}
The original Restormer Gaussian denoising setup already uses a standard multi-dataset recipe built from DIV2K~\cite{agustsson2017ntire}, Flickr2K~\cite{lim2017flickr2k}, BSD500~\cite{arbelaez2011bsd500}, and WED~\cite{ma2017wed}. We strengthen this setting by explicitly shifting the focus toward larger public image diversity. The final training dataset draws from seven public or officially released high-resolution image sources: DIV2K~\cite{agustsson2017ntire}, Flickr2K~\cite{lim2017flickr2k}, OST~\cite{wang2018recovering,wang2018esrgan}, LSDIR~\cite{li2023lsdir}, LIU4K-v2~\cite{liu2020liu4kv2}, the NKUSR8K release distributed with the DiT4SR project~\cite{duan2025nkusr8k}, and DIV8K~\cite{gu2019div8k}. In practice, the ultra-high-resolution sources are first cropped into approximately 2K sub-images before patch sampling so that the training distribution remains manageable while still benefiting from richer texture statistics.

Training is organized in two stages. We start from the public Restormer $\sigma = 50$ pretrained model rather than from random initialization. In Stage I, we continue training on a base expanded corpus consisting of DIV2K~\cite{agustsson2017ntire}, Flickr2K~\cite{lim2017flickr2k}, OST~\cite{wang2018recovering,wang2018esrgan}, and LSDIR~\cite{li2023lsdir}. This stage establishes a stronger and more stable restoration prior under increased scene and texture diversity. In Stage II, we further introduce LIU4K-v2~\cite{liu2020liu4kv2}, the NKUSR8K release distributed with the DiT4SR project~\cite{duan2025nkusr8k}, and DIV8K~\cite{gu2019div8k}, which enlarge the visual diversity and encourage broader generalization. Across both stages, the main point is not merely to use more data, but to obtain a stronger final checkpoint through a more capable data recipe. After combining the two stages, the overall training pool contains seven sources and 143,679 images.

To keep the presentation compact, we summarize the full recipe in one table instead of keeping separate dataset and optimizer tables. Both stages use AdamW and MSE loss on $4 \times$ NVIDIA H200 GPUs, while the learning rate drops from $1 \times 10^{-4}$ in Stage I to $1 \times 10^{-5}$ in Stage II.

\begin{table}[t]
    \centering
    \scriptsize
    \setlength{\tabcolsep}{3pt}
    \caption{Two-stage training summary.}
    \label{tab:train_summary}
    \begin{tabularx}{\linewidth}{L{0.18\linewidth}YY}
        \toprule
        Item & Stage I & Stage II \\
        \midrule
        Data & DIV2K / Flickr2K / OST / LSDIR & Stage I + LIU4K-v2 / NKUSR8K / DIV8K \\
        Patch & $256 \rightarrow 448 \rightarrow 768$ & 768 \\
        Batch & $4 \rightarrow 2 \rightarrow 1$ & 4 \\
        Iterations & 300K & 300K \\
        Initial LR & $1 \times 10^{-4}$ & $1 \times 10^{-5}$ \\
        Goal & Stable restoration prior & Broader texture diversity \\
        \bottomrule
    \end{tabularx}
\end{table}

\subsection{Test-Time Enhancement}
\paragraph{Geometric self-ensemble.}
Although a single forward pass already produces a strong result, neural denoisers are not strictly invariant to flips and transposes. This motivates $\times 8$ geometric self-ensemble, a test-time strategy explicitly used in strong restoration baselines such as SwinIR and SRFormer~\cite{liang2021swinir,zhou2023srformer}. Let $T_k(\cdot)$ denote the $k$-th geometric transform. The final prediction is
\begin{equation}
\hat{y} = \frac{1}{K}\sum_{k=1}^{K} T_k^{-1}\big(f_{\theta}(T_k(x))\big), \qquad K = 8.
\end{equation}
This strategy does not change the model parameters or training cost, but increases inference time by about a factor of eight. It is therefore a typical accuracy-through-computation trade-off.

\paragraph{TLC-style local inference wrapper.}
The final implementation also keeps a TLC-style local inference wrapper~\cite{chu2022improving}. Importantly, this wrapper is not treated as a new independently trained backbone. It is a deployment-level wrapper around the same Restormer weights, kept mainly for implementation consistency and local inference adaptation. The experiments later show that its measurable quantitative effect is negligible in the current Gaussian denoising setting, so it should not be interpreted as the dominant source of the final performance gain.

\begin{figure*}[t]
    \centering
    \includegraphics[width=0.96\textwidth]{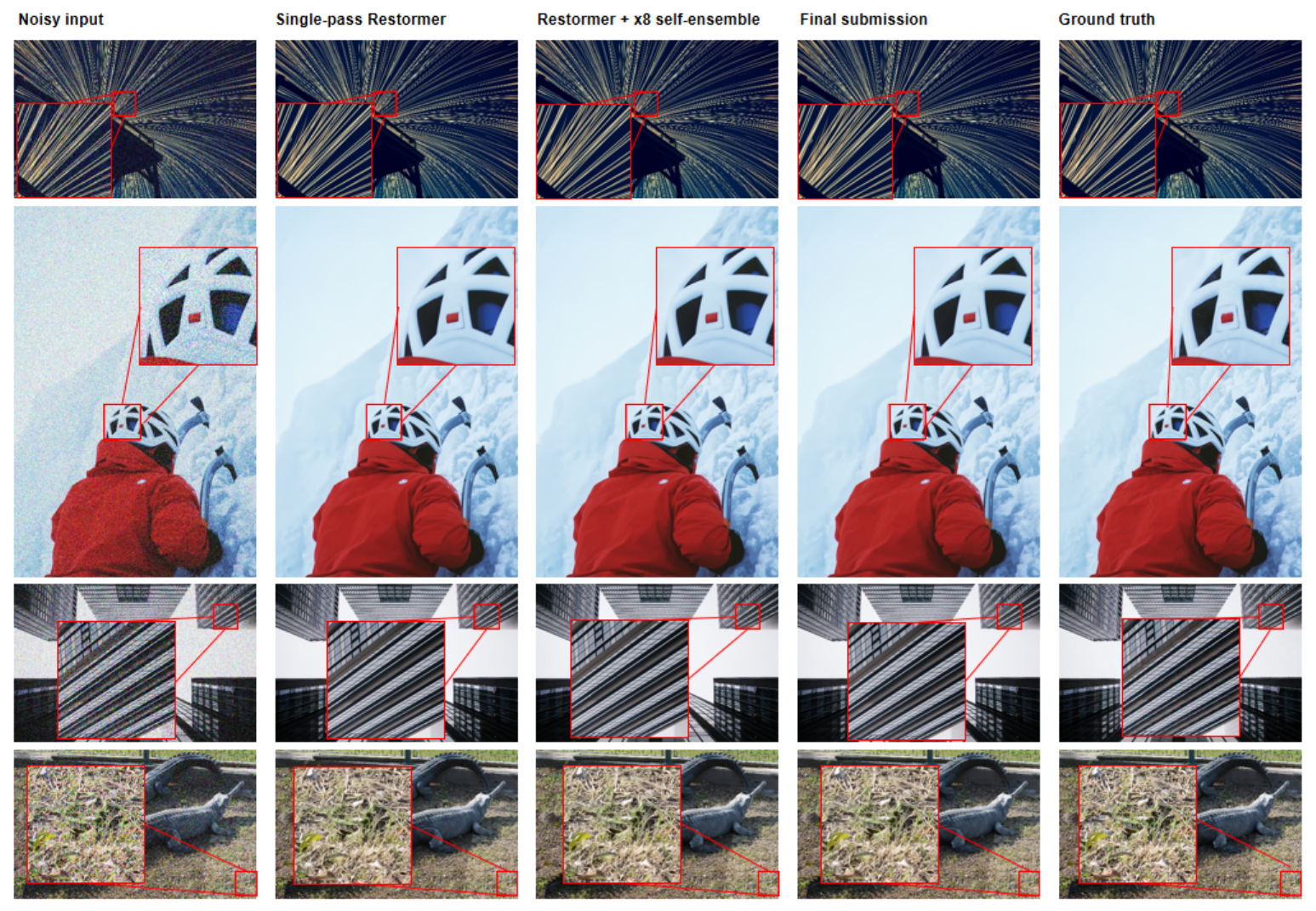}
    \caption{Qualitative comparison among the noisy input, single-pass Restormer, Restormer with $\times 8$ self-ensemble, final submission, and ground truth. The largest visual gains appear in line-rich, structured, and texture-dense regions.}
    \label{fig:qualitative}
\end{figure*}

\subsection{Final Inference Pipeline}
The final inference pipeline is straightforward. A three-channel noisy RGB input is first cropped to a spatial size divisible by eight. The image is then normalized to the $[0,1]$ range and converted to tensor form. The TLC-style wrapped Restormer backbone~\cite{zamir2022restormer,chu2022improving} restores the image, optionally under the eight geometric transforms used for self-ensemble. The predictions are mapped back to the RGB image space after inverse transformation and averaging. Clean reference images are only used for metric computation, not as inputs to the model.

\section{Experiments}
\subsection{Experimental Settings}
Training follows the expanded public corpus and two-stage strategy described in Section~3. Validation is performed on a unified 100-image protocol, and PSNR and SSIM are the main evaluation metrics. Unless otherwise stated, the main reported system is the TLC-style wrapped Restormer with $\times 8$ self-ensemble.

The experiments are organized to isolate the true source of improvement. We first report the main final result under the unified protocol. We then study the influence of self-ensemble and the TLC-style wrapper through inference ablation. After that, we compare checkpoints from different training stages to examine parameter evolution. Finally, we compare the resulting model with the public Restormer $\sigma = 50$ pretrained baseline under the same protocol, while the deployment cost is discussed directly from the same ablation table instead of a separate efficiency table.

\subsection{Main Result}
The final wrapped system with $\times 8$ self-ensemble reaches 30.7622 dB PSNR and 0.8607 SSIM, with an average inference time of 8815.73 ms and a peak memory usage of 36956 MB on the unified 100-image validation set. Because these final numbers are already covered again in the same-protocol baseline comparison, we report them here directly in text instead of keeping a separate one-row table. This confirms that the strengthened training-and-inference recipe consistently yields strong Gaussian denoising performance under one unified protocol.

\subsection{Inference Ablation}
Table~\ref{tab:ablation} restores the full four-way inference ablation. The most stable finding is that geometric self-ensemble provides a small but consistent gain regardless of whether the TLC-style wrapper is enabled. By contrast, the wrapper itself contributes almost nothing measurable in this task.

\begin{table}[t]
    \centering
    \scriptsize
    \setlength{\tabcolsep}{3pt}
    \caption{Inference ablation under the unified 100-image protocol.}
    \label{tab:ablation}
    \begin{tabularx}{\linewidth}{L{0.33\linewidth}*{4}{>{\centering\arraybackslash}X}}
        \toprule
        Variant & PSNR & SSIM & Time/ms & Mem/MB \\
        \midrule
        Restormer, 1-pass & 30.7349 & 0.8603 & 1063.01 & 36089 \\
        Restormer, $\times 8$ & 30.7622 & 0.8607 & 8759.55 & 36856 \\
        Wrapped, 1-pass & 30.7349 & 0.8603 & 1053.60 & 36190 \\
        Wrapped, $\times 8$ & 30.7622 & 0.8607 & 8815.73 & 36956 \\
        \bottomrule
    \end{tabularx}
\end{table}

The gain from self-ensemble is about 0.0273 dB PSNR and 0.0004 SSIM, which is small but stable. The wrapped and unwrapped variants are numerically almost identical. This supports the central interpretation of the paper: the dominant inference-side gain comes from self-ensemble, while the TLC-style wrapper is not the main reason for the final improvement.

\subsection{Checkpoint Evolution}
To understand how much of the improvement comes from the final training stage, we directly compare an earlier checkpoint with the final model in text rather than keeping a separate low-yield table. Intermediate checkpoint A (about 200K iterations) reaches 30.7580 dB PSNR and 0.8606 SSIM, while the final model reaches 30.7622 dB and 0.8607. The gap is therefore only about 0.0042 dB PSNR and 0.0001 SSIM, suggesting that the model had already nearly converged before the end of Stage II.

\subsection{Comparison with the Restormer Baseline}
\begin{figure}[t]
    \centering
    \includegraphics[width=\linewidth]{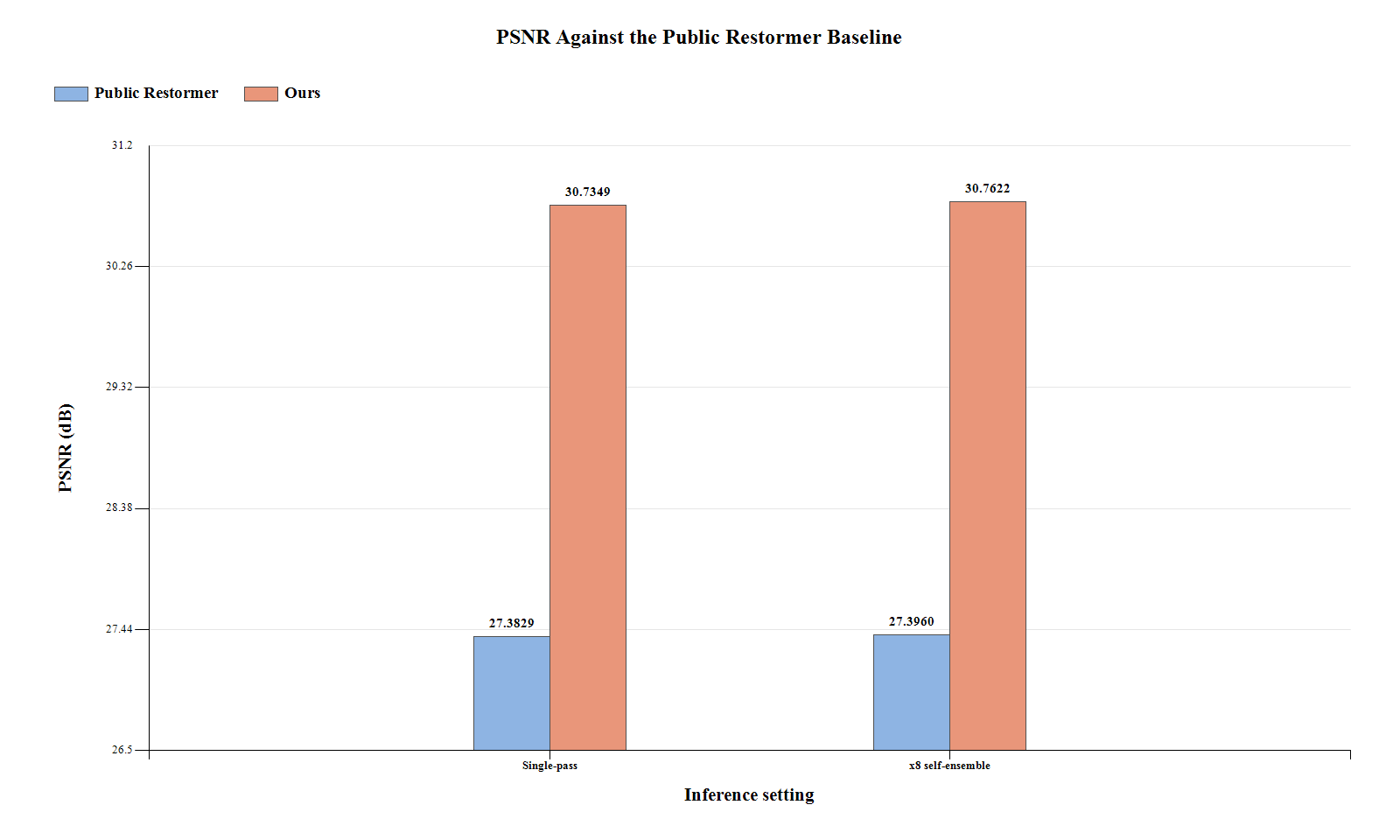}
    \caption{Same-protocol PSNR comparison with the public Restormer $\sigma = 50$ pretrained baseline. The gap remains large in both the single-pass and $\times 8$ self-ensemble settings; the corresponding SSIM gains are +0.0737 in both cases.}
    \label{fig:baseline}
\end{figure}

Figure~\ref{fig:baseline} converts the same-protocol baseline comparison into a compact visual summary. The performance gap is already large in the single-pass setting, and self-ensemble only adds a relatively small additional improvement on top of that stronger checkpoint. Quantitatively, the PSNR margin is +3.3520 dB for single-pass inference and +3.3662 dB with $\times 8$ self-ensemble, while the SSIM gain remains +0.0737 in both settings. This again suggests that the major gain comes from stronger checkpoints learned from the expanded training dataset rather than from the wrapper alone.

\subsection{Efficiency and Visual Discussion}
As shown in Table~\ref{tab:ablation}, on the wrapped path, $\times 8$ self-ensemble increases average inference time from 1053.60 ms to 8815.73 ms and raises peak memory from 36190 MB to 36956 MB. The unwrapped path shows the same pattern, confirming that the final gain is real but computation-heavy.

The qualitative observations are also image-dependent. As shown in Figure~\ref{fig:qualitative}, self-ensemble is most helpful in texture-rich and regions with regular structures. Sample 0828 shows a gain of 0.2463 dB PSNR in a high-frequency line region, and sample 0845 shows a gain of 0.1163 dB on a structured texture example. By contrast, the gains shrink noticeably in smoother or more ambiguous regions: samples 0844 and 0895 improve by only 0.0328 dB and 0.0084 dB, respectively. This indicates that self-ensemble should be viewed as a gradual robustness refinement rather than as a uniformly large structural improvement.

Taken together, Table~\ref{tab:ablation} and Figure~\ref{fig:baseline} and Figure~\ref{fig:qualitative} make the accuracy-efficiency trade-off explicit. If the goal is to obtain the strongest final restoration quality, keeping $\times 8$ self-ensemble is reasonable. If throughput or deployment cost matters more, single-pass inference already provides strong and stable restoration quality.

\section{Conclusion}
This paper presents a Restormer-based Gaussian color image denoising pipeline built around an expanded public training dataset and self-ensemble inference. Instead of redesigning the backbone, the method strengthens the final checkpoint through a data-centric two-stage recipe and then yields a small but consistent additional gain through $\times 8$ geometric self-ensemble. On the unified 100-image validation protocol, the final model reaches 30.7622 dB PSNR and 0.8607 SSIM, and improves over the public Restormer $\sigma = 50$ pretrained baseline by up to 3.3662 dB PSNR under the same protocol.

{\small
\nocite{*}
\bibliographystyle{plain}
\bibliography{references}

@article{dabov2007image,
  title={Image denoising by sparse 3-D transform-domain collaborative filtering},
  author={Dabov, Kostadin and Foi, Alessandro and Katkovnik, Vladimir and Egiazarian, Karen},
  journal={IEEE Transactions on Image Processing},
  volume={16},
  number={8},
  pages={2080--2095},
  year={2007},
  publisher={IEEE}
}

@inproceedings{buades2005non,
  title={A non-local algorithm for image denoising},
  author={Buades, Antoni and Coll, Bartomeu and Morel, J-M},
  booktitle={Proceedings of the 2005 IEEE Computer Society Conference on Computer Vision and Pattern Recognition (CVPR'05)},
  volume={2},
  pages={60--65},
  year={2005},
  organization={IEEE}
}

@article{elad2006image,
  title={Image denoising via sparse and redundant representations over learned dictionaries},
  author={Elad, Michael and Aharon, Michal},
  journal={IEEE Transactions on Image Processing},
  volume={15},
  number={12},
  pages={3736--3745},
  year={2006},
  publisher={IEEE}
}

@article{chen2016trainable,
  title={Trainable nonlinear reaction diffusion: A flexible framework for fast and effective image restoration},
  author={Chen, Yunjin and Pock, Thomas},
  journal={IEEE Transactions on Pattern Analysis and Machine Intelligence},
  volume={39},
  number={6},
  pages={1256--1272},
  year={2016},
  publisher={IEEE}
}

@article{zhang2017beyond,
  title={Beyond a Gaussian denoiser: Residual learning of deep {CNN} for image denoising},
  author={Zhang, Kai and Zuo, Wangmeng and Chen, Yunjin and Meng, Deyu and Zhang, Lei},
  journal={IEEE Transactions on Image Processing},
  volume={26},
  number={7},
  pages={3142--3155},
  year={2017},
  publisher={IEEE}
}

@article{zhang2018ffdnet,
  title={{FFDNet}: Toward a fast and flexible solution for {CNN}-based image denoising},
  author={Zhang, Kai and Zuo, Wangmeng and Zhang, Lei},
  journal={IEEE Transactions on Image Processing},
  volume={27},
  number={9},
  pages={4608--4622},
  year={2018},
  publisher={IEEE}
}

@inproceedings{anwar2019real,
  title={Real image denoising with feature attention},
  author={Anwar, Saeed and Barnes, Nick},
  booktitle={Proceedings of the IEEE/CVF International Conference on Computer Vision},
  pages={3155--3164},
  year={2019}
}

@inproceedings{guo2019toward,
  title={Toward convolutional blind denoising of real photographs},
  author={Guo, Shi and Yan, Zifei and Zhang, Kai and Zuo, Wangmeng and Zhang, Lei},
  booktitle={Proceedings of the IEEE/CVF Conference on Computer Vision and Pattern Recognition},
  pages={1712--1722},
  year={2019}
}

@inproceedings{zamir2020cycleisp,
  title={{CycleISP}: Real image restoration via improved data synthesis},
  author={Zamir, Syed Waqas and Arora, Aditya and Khan, Salman and Hayat, Munawar and Khan, Fahad Shahbaz and Yang, Ming-Hsuan and Shao, Ling},
  booktitle={Proceedings of the IEEE/CVF Conference on Computer Vision and Pattern Recognition},
  pages={2696--2705},
  year={2020}
}

@inproceedings{zamir2020learning,
  title={Learning enriched features for real image restoration and enhancement},
  author={Zamir, Syed Waqas and Arora, Aditya and Khan, Salman and Hayat, Munawar and Khan, Fahad Shahbaz and Yang, Ming-Hsuan and Shao, Ling},
  booktitle={Proceedings of the European Conference on Computer Vision (ECCV)},
  pages={492--511},
  year={2020},
  organization={Springer}
}

@article{zamir2022learning,
  title={Learning enriched features for fast image restoration and enhancement},
  author={Zamir, Syed Waqas and Arora, Aditya and Khan, Salman and Hayat, Munawar and Khan, Fahad Shahbaz and Yang, Ming-Hsuan and Shao, Ling},
  journal={IEEE Transactions on Pattern Analysis and Machine Intelligence},
  volume={45},
  number={2},
  pages={1934--1948},
  year={2022},
  publisher={IEEE}
}

@inproceedings{lehtinen2018noise2noise,
  title={{Noise2Noise}: Learning image restoration without clean data},
  author={Lehtinen, Jaakko and Munkberg, Jacob and Hasselgren, Jon and Laine, Samuli and Karras, Tero and Aittala, Miika and Aila, Timo},
  booktitle={Proceedings of the 35th International Conference on Machine Learning},
  pages={2965--2974},
  year={2018},
  volume={80},
  series={Proceedings of Machine Learning Research},
  publisher={PMLR}
}

@inproceedings{krull2019noise2void,
  title={{Noise2Void}: Learning denoising from single noisy images},
  author={Krull, Alexander and Buchholz, Tim-Oliver and Jug, Florian},
  booktitle={Proceedings of the IEEE/CVF Conference on Computer Vision and Pattern Recognition},
  pages={2129--2137},
  year={2019}
}

@inproceedings{batson2019noise2self,
  title={{Noise2Self}: Blind denoising by self-supervision},
  author={Batson, Joshua and Royer, Loic},
  booktitle={Proceedings of the International Conference on Machine Learning},
  pages={524--533},
  year={2019},
  organization={PMLR}
}

@inproceedings{quan2020self2self,
  title={{Self2Self} with dropout: Learning self-supervised denoising from single image},
  author={Quan, Yuhui and Chen, Mingqin and Pang, Tongyao and Ji, Hui},
  booktitle={Proceedings of the IEEE/CVF Conference on Computer Vision and Pattern Recognition},
  pages={1890--1898},
  year={2020}
}

@inproceedings{huang2021neighbor2neighbor,
  title={{Neighbor2Neighbor}: Self-supervised denoising from single noisy images},
  author={Huang, Tao and Li, Songjiang and Jia, Xu and Lu, Huchuan and Liu, Jianzhuang},
  booktitle={Proceedings of the IEEE/CVF Conference on Computer Vision and Pattern Recognition},
  pages={14781--14790},
  year={2021}
}

@article{laine2019high,
  title={High-quality self-supervised deep image denoising},
  author={Laine, Samuli and Karras, Tero and Lehtinen, Jaakko and Aila, Timo},
  journal={Advances in Neural Information Processing Systems},
  volume={32},
  year={2019}
}

@inproceedings{tassano2020fastdvdnet,
  title={{FastDVDnet}: Towards real-time deep video denoising without flow estimation},
  author={Tassano, Matias and Delon, Julie and Veit, Thomas},
  booktitle={Proceedings of the IEEE/CVF Conference on Computer Vision and Pattern Recognition},
  pages={1354--1363},
  year={2020}
}

@inproceedings{zamir2021multi,
  title={Multi-stage progressive image restoration},
  author={Zamir, Syed Waqas and Arora, Aditya and Khan, Salman and Hayat, Munawar and Khan, Fahad Shahbaz and Yang, Ming-Hsuan and Shao, Ling},
  booktitle={Proceedings of the IEEE/CVF Conference on Computer Vision and Pattern Recognition},
  pages={14821--14831},
  year={2021}
}

@inproceedings{chen2021pre,
  title={Pre-trained image processing transformer},
  author={Chen, Hanting and Wang, Yunhe and Guo, Tianyu and Xu, Chang and Deng, Yiping and Liu, Zhenhua and Ma, Siwei and Xu, Chunjing and Xu, Chao and Gao, Wen},
  booktitle={Proceedings of the IEEE/CVF Conference on Computer Vision and Pattern Recognition},
  pages={12299--12310},
  year={2021}
}

@inproceedings{liu2021swin,
  title={{Swin Transformer}: Hierarchical vision transformer using shifted windows},
  author={Liu, Ze and Lin, Yutong and Cao, Yue and Hu, Han and Wei, Yixuan and Zhang, Zheng and Lin, Stephen and Guo, Baining},
  booktitle={Proceedings of the IEEE/CVF International Conference on Computer Vision},
  pages={10012--10022},
  year={2021}
}

@inproceedings{liang2021swinir,
  title={{SwinIR}: Image restoration using {Swin Transformer}},
  author={Liang, Jingyun and Cao, Jiezhang and Sun, Guolei and Zhang, Kai and Van Gool, Luc and Timofte, Radu},
  booktitle={Proceedings of the IEEE/CVF International Conference on Computer Vision},
  pages={1833--1844},
  year={2021}
}

@inproceedings{wang2022uformer,
  title={{Uformer}: A general U-shaped transformer for image restoration},
  author={Wang, Zhendong and Cun, Xiaodong and Bao, Jianmin and Zhou, Wengang and Liu, Jianzhuang and Li, Houqiang},
  booktitle={Proceedings of the IEEE/CVF Conference on Computer Vision and Pattern Recognition},
  pages={17683--17693},
  year={2022}
}

@inproceedings{zamir2022restormer,
  title={{Restormer}: Efficient transformer for high-resolution image restoration},
  author={Zamir, Syed Waqas and Arora, Aditya and Khan, Salman and Hayat, Munawar and Khan, Fahad Shahbaz and Yang, Ming-Hsuan},
  booktitle={Proceedings of the IEEE/CVF Conference on Computer Vision and Pattern Recognition},
  pages={5728--5739},
  year={2022}
}

@inproceedings{chen2021hinet,
  title={{HiNet}: Half instance normalization network for image restoration},
  author={Chen, Liangyu and Lu, Xin and Zhang, Jie and Chu, Xiaojie and Chen, Chengpeng},
  booktitle={Proceedings of the IEEE/CVF Conference on Computer Vision and Pattern Recognition},
  pages={182--192},
  year={2021}
}

@inproceedings{chen2022simple,
  title={Simple baselines for image restoration},
  author={Chen, Liangyu and Chu, Xiaojie and Zhang, Xiangyu and Sun, Jian},
  booktitle={Proceedings of the European Conference on Computer Vision (ECCV)},
  pages={17--33},
  year={2022},
  organization={Springer}
}

@inproceedings{tu2022maxim,
  title={{MAXIM}: Multi-axis {MLP} for image processing},
  author={Tu, Zhengzhong and Talebi, Hossein and Zhang, Han and Yang, Feng and Milanfar, Peyman and Bovik, Alan and Li, Yinxiao},
  booktitle={Proceedings of the IEEE/CVF Conference on Computer Vision and Pattern Recognition},
  pages={5769--5780},
  year={2022}
}

@inproceedings{guo2024mambair,
  title={{MambaIR}: A simple baseline for image restoration with state-space model},
  author={Guo, Hang and Li, Jinmin and Dai, Tao and Ouyang, Zhihao and Ren, Xudong and Xia, Shu-Tao},
  booktitle={Proceedings of the European Conference on Computer Vision (ECCV)},
  pages={222--241},
  year={2024},
  organization={Springer}
}

@inproceedings{guo2025mambairv2,
  title={{MambaIRv2}: Attentive state space restoration},
  author={Guo, Hang and Guo, Yong and Zha, Yaohua and Zhang, Yulun and Li, Wenbo and Dai, Tao and Xia, Shu-Tao and Li, Yawei},
  booktitle={Proceedings of the IEEE/CVF Conference on Computer Vision and Pattern Recognition},
  pages={28124--28133},
  year={2025}
}

@inproceedings{chu2022improving,
  title={Improving image restoration by revisiting global information aggregation},
  author={Chu, Xiaojie and Chen, Liangyu and Chen, Chengpeng and Lu, Xin},
  booktitle={Proceedings of the European Conference on Computer Vision (ECCV)},
  pages={53--71},
  year={2022},
  organization={Springer}
}

@article{chen2022cross,
  title={Cross aggregation transformer for image restoration},
  author={Chen, Zheng and Zhang, Yulun and Gu, Jinjin and Kong, Linghe and Yuan, Xin and others},
  journal={Advances in Neural Information Processing Systems},
  volume={35},
  pages={25478--25490},
  year={2022}
}

@inproceedings{agustsson2017ntire,
  title={{NTIRE} 2017 challenge on single image super-resolution: Dataset and study},
  author={Agustsson, Eirikur and Timofte, Radu},
  booktitle={Proceedings of the IEEE Conference on Computer Vision and Pattern Recognition Workshops},
  pages={126--135},
  year={2017}
}

@inproceedings{wang2018recovering,
  title={Recovering realistic texture in image super-resolution by deep spatial feature transform},
  author={Wang, Xintao and Yu, Ke and Dong, Chao and Loy, Chen Change},
  booktitle={Proceedings of the IEEE Conference on Computer Vision and Pattern Recognition},
  pages={606--615},
  year={2018}
}

@inproceedings{wang2018esrgan,
  title={{ESRGAN}: Enhanced super-resolution generative adversarial networks},
  author={Wang, Xintao and Yu, Ke and Wu, Shixiang and Gu, Jinjin and Liu, Yihao and Dong, Chao and Qiao, Yu and Loy, Chen Change},
  booktitle={Proceedings of the European Conference on Computer Vision Workshops},
  year={2018}
}

@inproceedings{li2023lsdir,
  title={{LSDIR}: A large-scale dataset for image restoration},
  author={Li, Yawei and Zhang, Kai and Liang, Jingyun and Cao, Jiezhang and Liu, Ce and Gong, Rui and Zhang, Yulun and Tang, Hao and Liu, Yun and Demandolx, Denis and others},
  booktitle={Proceedings of the IEEE/CVF Conference on Computer Vision and Pattern Recognition},
  pages={1775--1787},
  year={2023}
}

@article{anaya2018renoir,
  title={{RENOIR}--A dataset for real low-light image noise reduction},
  author={Anaya, Josue and Barbu, Adrian},
  journal={Journal of Visual Communication and Image Representation},
  volume={51},
  pages={144--154},
  year={2018},
  publisher={Elsevier}
}

@article{chen2022real,
  title={Real-world single image super-resolution: A brief review},
  author={Chen, Honggang and He, Xiaohai and Qing, Linbo and Wu, Yuanyuan and Ren, Chao and Sheriff, Ray E and Zhu, Ce},
  journal={Information Fusion},
  volume={79},
  pages={124--145},
  year={2022},
  publisher={Elsevier}
}

@article{liu2022blind,
  title={Blind image super-resolution: A survey and beyond},
  author={Liu, Anran and Liu, Yihao and Gu, Jinjin and Qiao, Yu and Dong, Chao},
  journal={IEEE Transactions on Pattern Analysis and Machine Intelligence},
  volume={45},
  number={5},
  pages={5461--5480},
  year={2022},
  publisher={IEEE}
}

@inproceedings{zhou2023srformer,
  title={{SRFormer}: Permuted self-attention for single image super-resolution},
  author={Zhou, Yupeng and Li, Zhen and Guo, Chun-Le and Bai, Song and Cheng, Ming-Ming and Hou, Qibin},
  booktitle={Proceedings of the IEEE/CVF International Conference on Computer Vision},
  pages={12780--12791},
  year={2023}
}

@inproceedings{chen2023activating,
  title={Activating more pixels in image super-resolution transformer},
  author={Chen, Xiangyu and Wang, Xintao and Zhou, Jiantao and Qiao, Yu and Dong, Chao},
  booktitle={Proceedings of the IEEE/CVF Conference on Computer Vision and Pattern Recognition},
  pages={22367--22377},
  year={2023}
}

@inproceedings{zhang2022efficient,
  title={Efficient long-range attention network for image super-resolution},
  author={Zhang, Xindong and Zeng, Hui and Guo, Shi and Zhang, Lei},
  booktitle={Proceedings of the European Conference on Computer Vision (ECCV)},
  pages={649--667},
  year={2022},
  organization={Springer}
}

@article{zhang2021plug,
  title={Plug-and-play image restoration with deep denoiser prior},
  author={Zhang, Kai and Li, Yawei and Zuo, Wangmeng and Zhang, Lei and Van Gool, Luc and Timofte, Radu},
  journal={IEEE Transactions on Pattern Analysis and Machine Intelligence},
  volume={44},
  number={10},
  pages={6360--6376},
  year={2021},
  publisher={IEEE}
}

@inproceedings{brooks2019unprocessing,
  title={Unprocessing images for learned raw denoising},
  author={Brooks, Tim and Mildenhall, Ben and Xue, Tianfan and Chen, Jiawen and Sharlet, Dillon and Barron, Jonathan T},
  booktitle={Proceedings of the IEEE/CVF Conference on Computer Vision and Pattern Recognition},
  pages={11036--11045},
  year={2019}
}

@inproceedings{tai2017memnet,
  title={{MemNet}: A persistent memory network for image restoration},
  author={Tai, Ying and Yang, Jian and Liu, Xiaoming and Xu, Chunyan},
  booktitle={Proceedings of the IEEE International Conference on Computer Vision},
  pages={4539--4547},
  year={2017}
}

@article{zhang2023practical,
  title={Practical blind image denoising via {Swin-Conv-UNet} and data synthesis},
  author={Zhang, Kai and Li, Yawei and Liang, Jingyun and Cao, Jiezhang and Zhang, Yulun and Tang, Hao and Fan, Deng-Ping and Timofte, Radu and Gool, Luc Van},
  journal={Machine Intelligence Research},
  volume={20},
  number={6},
  pages={822--836},
  year={2023},
  publisher={Springer}
}

@inproceedings{godard2018deep,
  title={Deep burst denoising},
  author={Godard, Cl{\'e}ment and Matzen, Kevin and Uyttendaele, Matt},
  booktitle={Proceedings of the European Conference on Computer Vision (ECCV)},
  pages={538--554},
  year={2018}
}

@inproceedings{claus2019videnn,
  title={{ViDeNN}: Deep blind video denoising},
  author={Claus, Michele and Van Gemert, Jan},
  booktitle={Proceedings of the IEEE/CVF Conference on Computer Vision and Pattern Recognition Workshops},
  pages={0--0},
  year={2019}
}

@inproceedings{nam2016holistic,
  title={A holistic approach to cross-channel image noise modeling and its application to image denoising},
  author={Nam, Seonghyeon and Hwang, Youngbae and Matsushita, Yasuyuki and Kim, Seon Joo},
  booktitle={Proceedings of the IEEE Conference on Computer Vision and Pattern Recognition},
  pages={1683--1691},
  year={2016}
}

@inproceedings{abdelhamed2019noise,
  title={{Noise Flow}: Noise modeling with conditional normalizing flows},
  author={Abdelhamed, Abdelrahman and Brubaker, Marcus A and Brown, Michael S},
  booktitle={Proceedings of the IEEE/CVF International Conference on Computer Vision},
  pages={3165--3173},
  year={2019}
}

@article{chen2019real,
  title={Real-world image denoising with deep boosting},
  author={Chen, Chang and Xiong, Zhiwei and Tian, Xinmei and Zha, Zheng-Jun and Wu, Feng},
  journal={IEEE Transactions on Pattern Analysis and Machine Intelligence},
  volume={42},
  number={12},
  pages={3071--3087},
  year={2019},
  publisher={IEEE}
}

@inproceedings{abdelhamed2018high,
  title={A high-quality denoising dataset for smartphone cameras},
  author={Abdelhamed, Abdelrahman and Lin, Stephen and Brown, Michael S},
  booktitle={Proceedings of the IEEE Conference on Computer Vision and Pattern Recognition},
  pages={1692--1700},
  year={2018}
}

@inproceedings{plotz2017benchmarking,
  title={Benchmarking denoising algorithms with real photographs},
  author={Plotz, Tobias and Roth, Stefan},
  booktitle={Proceedings of the IEEE Conference on Computer Vision and Pattern Recognition},
  pages={1586--1595},
  year={2017}
}

@inproceedings{abdelhamed2019ntire,
  title={{NTIRE} 2019 challenge on real image denoising: Methods and results},
  author={Abdelhamed, Abdelrahman and Timofte, Radu and Brown, Michael S. and others},
  booktitle={Proceedings of the IEEE/CVF Conference on Computer Vision and Pattern Recognition Workshops},
  pages={2197--2210},
  year={2019},
  doi={10.1109/CVPRW.2019.00273}
}

@inproceedings{abdelhamed2020ntire,
  title={{NTIRE} 2020 challenge on real image denoising: Dataset, methods and results},
  author={Abdelhamed, Abdelrahman and Afifi, Mahmoud and Timofte, Radu and Brown, Michael S. and others},
  booktitle={Proceedings of the IEEE/CVF Conference on Computer Vision and Pattern Recognition Workshops},
  pages={2077--2088},
  year={2020},
  doi={10.1109/CVPRW50498.2020.00256}
}

@inproceedings{li2023ntire,
  title={NTIRE 2023 challenge on image denoising: Methods and results},
  author={Li, Yawei and Zhang, Yulun and Timofte, Radu and Van Gool, Luc and Tu, Zhijun and Du, Kunpeng and Wang, Hailing and Chen, Hanting and Li, Wei and Wang, Xiaofei and others},
  booktitle={Proceedings of the IEEE/CVF Conference on Computer Vision and Pattern Recognition},
  pages={1905--1921},
  year={2023}
}

@misc{lim2017flickr2k,
  title={{Flickr2K} dataset},
  author={Lim, Bee and Son, Sanghyun and Kim, Heewon and Nah, Seungjun and Lee, Kyoung Mu},
  year={2017},
  howpublished={Official dataset release accompanying the {NTIRE2017}/{EDSR} repository},
  note={Dataset collected by the authors using the Flickr API},
  url={https://github.com/limbee/NTIRE2017}
}

@article{arbelaez2011bsd500,
  title={Contour detection and hierarchical image segmentation},
  author={Arbel{\'a}ez, Pablo and Maire, Michael and Fowlkes, Charless and Malik, Jitendra},
  journal={IEEE Transactions on Pattern Analysis and Machine Intelligence},
  volume={33},
  number={5},
  pages={898--916},
  year={2011},
  publisher={IEEE}
}

@article{ma2017wed,
  title={{Waterloo Exploration Database}: New challenges for image quality assessment models},
  author={Ma, Kede and Duanmu, Zhengfang and Wu, Qingbo and Wang, Zhou and Yong, Hongwei and Li, Hongliang and Zhang, Lei},
  journal={IEEE Transactions on Image Processing},
  volume={26},
  number={2},
  pages={1004--1016},
  year={2017},
  publisher={IEEE}
}

@misc{liu2020liu4kv2,
  title={{LIU4K-v2} dataset},
  author={Liu, Jiaying and Liu, Dong and Yang, Wenhan and Xia, Sifeng and Zhang, Xiaoshuai and Dai, Yuanying},
  year={2020},
  howpublished={Official dataset page},
  note={The official {LIU4K-v2} page recommends citing the accompanying compression artifact reduction benchmark paper},
  url={https://structpku.github.io/LIU4K_Dataset/LIU4K_v2.html}
}

@inproceedings{gu2019div8k,
  title={{DIV8K}: {DIVerse} 8K resolution image dataset},
  author={Gu, Shuhang and Lugmayr, Andreas and Danelljan, Martin and Fritsche, Manuel and Lamour, Julien and Timofte, Radu},
  booktitle={Proceedings of the IEEE/CVF International Conference on Computer Vision Workshops},
  pages={3512--3516},
  year={2019},
  publisher={IEEE}
}

@misc{duan2025nkusr8k,
  title={{NKUSR8K}: dataset release in the official {DiT4SR} project repository},
  author={Duan, Zheng-Peng and Zhang, Jiawei and Jin, Xin and Zhang, Ziheng and Xiong, Zheng and Zou, Dongqing and Ren, Jimmy S. and Guo, Chun-Le and Li, Chongyi},
  year={2025},
  howpublished={Official project repository},
  note={Repository documentation states that the {NKUSR8K} dataset is released for training with the {DiT4SR} project},
  url={https://github.com/Adam-duan/DiT4SR}
}

@article{liuderain2025,
    author={Liu, Shuhong and Chen, Xiang and Chen, Hongming and Xu, Quanfeng and Li, Mingrui},
    title={DeRainGS: Gaussian Splatting for Enhanced Scene Reconstruction in Rainy Environments},
    volume={39},
    DOI={10.1609/aaai.v39i5.32592},
    number={5},
    journal={Proceedings of the AAAI Conference on Artificial Intelligence},
    year={2025},
    pages={5558-5566}
}

@inproceedings{liu2025i2nerf,
    title={I2-NeRF: Learning Neural Radiance Fields Under Physically-Grounded Media Interactions},
    author={Liu, Shuhong and Gu, Lin and Cui, Ziteng and Chu, Xuangeng and Harada, Tatsuya},
    booktitle={Advances in Neural Information Processing Systems},
    year={2025},
}

@inproceedings{lisgs2025,
    title={SGS-SLAM: Semantic Gaussian Splatting for Neural Dense SLAM},
    author={Li, Mingrui and Liu, Shuhong and Zhou, Heng and Zhu, Guohao and Cheng, Na and Deng, Tianchen and Wang, Hongyu},
    booktitle={European Conference on Computer Vision},
    year={2025},
    pages={163--179},
}

@article{liumg2025,
    title={MG-SLAM: Structure Gaussian Splatting SLAM With Manhattan World Hypothesis}, 
    author={Liu, Shuhong and Deng, Tianchen and Zhou, Heng and Li, Liuzhuozheng and Wang, Hongyu and Wang, Danwei and Li, Mingrui},
    journal={IEEE Transactions on Automation Science and Engineering}, 
    year={2025},
    volume={22},
    number={},
    pages={17034-17049},
    doi={10.1109/TASE.2025.3575772}
}

@article{lidense2025,
    title={DenseSplat: Densifying Gaussian Splatting SLAM with Neural Radiance Prior},
    author={Li, Mingrui and Liu, Shuhong and Deng, Tianchen and Wang, Hongyu},
    journal={IEEE Transactions on Visualization \& Computer Graphics},
    year={2025},
    volume={},
    number={01},
    ISSN={1941-0506},
    pages={1-14},
    doi={10.1109/TVCG.2025.3617961},
    publisher={IEEE Computer Society},
}

@article{liu2025realx3d,
    title={RealX3D: A Physically-Degraded 3D Benchmark for Multi-view Visual Restoration and Reconstruction},
    author={Liu, Shuhong and Bao, Chenyu and Cui, Ziteng and Liu, Yun and Chu, Xuangeng and Gu, Lin and Conde, Marcos V and Umagami, Ryo and Hashimoto, Tomohiro and Hu, Zijian and others},
    journal={arXiv preprint arXiv:2512.23437},
    year={2026}
}

@article{liudenoise2026,
    title={Denoising the Deep Sky: Physics-Based CCD Noise Formation for Astronomical Imaging},
    author={Liu, Shuhong and Ge, Xining and Gu, Ziying and Gu, Lin and Cui, Ziteng and Chu, Xuangeng and Liu, Jun and Li, Dong and Harada, Tatsuya},
    journal={arXiv preprint arXiv:2601.23276},
    year={2026}
}

@article{cui2026unifying,
  title={Unifying Color and Lightness Correction with View-Adaptive Curve Adjustment for Robust 3D Novel View Synthesis},
  author={Cui, Ziteng and Liu, Shuhong and Dong, Xiaoyu and Chu, Xuangeng and Gu, Lin and Yang, Ming-Hsuan and Harada, Tatsuya},
  journal={arXiv preprint arXiv:2602.18322},
  year={2026}
}

@article{zhou2024mod,
  title={Mod-slam: Monocular dense mapping for unbounded 3d scene reconstruction},
  author={Zhou, Heng and Guo, Zhetao and Ren, Yuxiang and Liu, Shuhong and Zhang, Lechen and Zhang, Kaidi and Li, Mingrui},
  journal={IEEE Robotics and Automation Letters},
  volume={10},
  number={1},
  pages={484--491},
  year={2024},
  publisher={IEEE}
}

@inproceedings{chen2026sr4,
  title={The Fourth Challenge on Image Super-Resolution ($\times 4$) at {NTIRE} 2026: Benchmark Results and Method Overview},
  author={Chen, Zheng and Liu, Kai and Wang, Jingkai and Yan, Xianglong and Li, Jianze and Zhang, Ziqing and Gong, Jue and Li, Jiatong and Sun, Lei and Liu, Xiaoyang and Timofte, Radu and Zhang, Yulun and others},
  booktitle={Proceedings of the Computer Vision and Pattern Recognition Conference Workshops},
  year={2026}
}

@article{ge2026clip,
  title={CLIP-Guided Data Augmentation for Night-Time Image Dehazing},
  author={Ge, Xining and Yuan, Weijun and Chang, Gengjia and Li, Xuyang and Liu, Shuhong},
  journal={arXiv preprint arXiv:2604.05500},
  year={2026}
}

@article{chang2026training,
  title={Training-Free Model Ensemble for Single-Image Super-Resolution via Strong-Branch Compensation},
  author={Chang, Gengjia and Ge, Xining and Yuan, Weijun and Li, Zhan and Song, Qiurong and Zhu, Luen and Liu, Shuhong},
  journal={arXiv preprint arXiv:2604.11564},
  year={2026}
}

@article{ge2026dual,
  title={Dual-Branch Remote Sensing Infrared Image Super-Resolution},
  author={Ge, Xining and Chang, Gengjia and Yuan, Weijun and Li, Zhan and Chen, Zhanglu and Yao, Boyang and Chen, Yihang and Deng, Yifan and Liu, Shuhong},
  journal={arXiv preprint arXiv:2604.10112},
  year={2026}
}

@article{zheng20263d,
  title={{3D} Smoke Scene Reconstruction Guided by Vision Priors from Multimodal Large Language Models},
  author={Zheng, Xinye and Wang, Fei and Nie, Yiqi and Li, Kun and Chen, Junjie and Zhao, Jiaqi and Wei, Yanyan and Wu, Zhiliang},
  journal={arXiv preprint arXiv:2604.05687},
  year={2026}
}

@article{liu2026elog,
  title={{ELoG-GS}: Dual-Branch Gaussian Splatting with Luminance-Guided Enhancement for Extreme Low-light {3D} Reconstruction},
  author={Liu, Yuhao and Wang, Dingju and Zheng, Ziyang},
  journal={arXiv preprint arXiv:2604.12592},
  year={2026}
}

@article{fu2026smokegs,
  title={{SmokeGS-R}: Physics-Guided Pseudo-Clean {3DGS} for Real-World Multi-View Smoke Restoration},
  author={Fu, Xueming and Han, Lixia},
  journal={arXiv preprint arXiv:2604.05301},
  year={2026}
}

@article{cao2026gensmoke,
  title={{GenSmoke-GS}: A Multi-Stage Method for Novel View Synthesis from Smoke-Degraded Images Using a Generative Model},
  author={Cao, Qida and Hu, Xinyuan and Shi, Changyue and Ding, Jiajun and Yu, Zhou and Yu, Jun},
  journal={arXiv preprint arXiv:2604.03039},
  year={2026}
}

@article{zhu2026naka,
  title={{Naka-GS}: A Bionics-Inspired Dual-Branch Naka Correction and Progressive Point Pruning for Low-Light {3DGS}},
  author={Zhu, Runyu and Dong, SiXun and Zhang, Zhiqiang and Ye, Qingxia and Xu, Zhihua},
  journal={arXiv preprint arXiv:2604.11142},
  year={2026}
}

@article{guo2026reliability,
  title={Reliability-Aware Staged Low-Light Gaussian Splatting},
  author={Guo, Haojie and Xian, Ke},
  journal={ResearchGate preprint},
  year={2026}
}

@article{chen2026dehaze,
  title={Dehaze-then-Splat: Generative Dehazing with Physics-Informed {3D} Gaussian Splatting for Smoke-Free Novel View Synthesis},
  author={Chen, Boss and Wang, Hanqing},
  journal={arXiv preprint arXiv:2604.13589},
  year={2026}
}

@article{liu20263drr,
  title={{NTIRE} 2026 {3D} Restoration and Reconstruction in Adverse Conditions: {RealX3D} Challenge Results},
  author={Liu, Shuhong and Bao, Chenyu and Cui, Ziteng and Chu, Xuangeng and Ren, Bin and Gu, Lin and Chen, Xiang and Li, Mingrui and Ma, Long and Conde, Marcos V. and Timofte, Radu and others},
  journal={arXiv preprint arXiv:2604.04135},
  year={2026}
}

@article{ren2026esr,
  title={The Eleventh {NTIRE} 2026 Efficient Super-Resolution Challenge Report},
  author={Ren, Bin and Guo, Hang and Shu, Yan and Ma, Jiaqi and Cui, Ziteng and Liu, Shuhong and Mei, Guofeng and Sun, Lei and Wu, Zongwei and Khan, Fahad Shahbaz and Khan, Salman and Timofte, Radu and Li, Yawei and others},
  journal={arXiv preprint arXiv:2604.03198},
  year={2026}
}

@inproceedings{ntire26denoising, 
    title={{The Third Challenge on Image Denoising at NTIRE 2026: Methods and Results}}, 
    author={Sun, Lei and  Guo, Hang and  Ren, Bin and  Su, Shaolin and  Wang, Xian and  Pani Paudel, Danda and  Van Gool, Luc and  Timofte, Radu and  Li, Yawei and others},   
    booktitle={Proceedings of the IEEE/CVF Conference on Computer Vision and Pattern Recognition (CVPR) Workshops},  
    year = {2026} 
}
}

\end{document}